\def\year{2020}\relax
\documentclass[letterpaper]{article} 
\usepackage{aaai20}  
\usepackage{times}  
\usepackage{helvet} 
\usepackage{courier}  
\usepackage{caption}
\usepackage{blindtext}
\usepackage[hyphens]{url}  
\usepackage{graphicx} 
\usepackage{graphicx}  
\usepackage{colortbl}
\usepackage{xcolor}
\usepackage{multirow}
\usepackage{tabulary}
\usepackage{bm}
\usepackage{amsmath}

\title{Universal Person Re-Identification}
\author{Xu Lan\\
	{Queen Mary University of London}\\
	x.lan@qmul.ac.uk
	\And
	Xiatian Zhu\\
	{Vision Semantics Ltd.}\\
	eddy.zhuxt@gmail.com
	\And
	Shaogang Gong\\
	{Queen Mary University of London}\\
	s.gong@qmul.ac.uk}

\begin{document}
\twocolumn[{%
	\renewcommand\twocolumn[1][]{#1}%
	\maketitle
	\begin{center}
		\centering
		\vspace{-1cm}
		  \includegraphics[width=\textwidth]{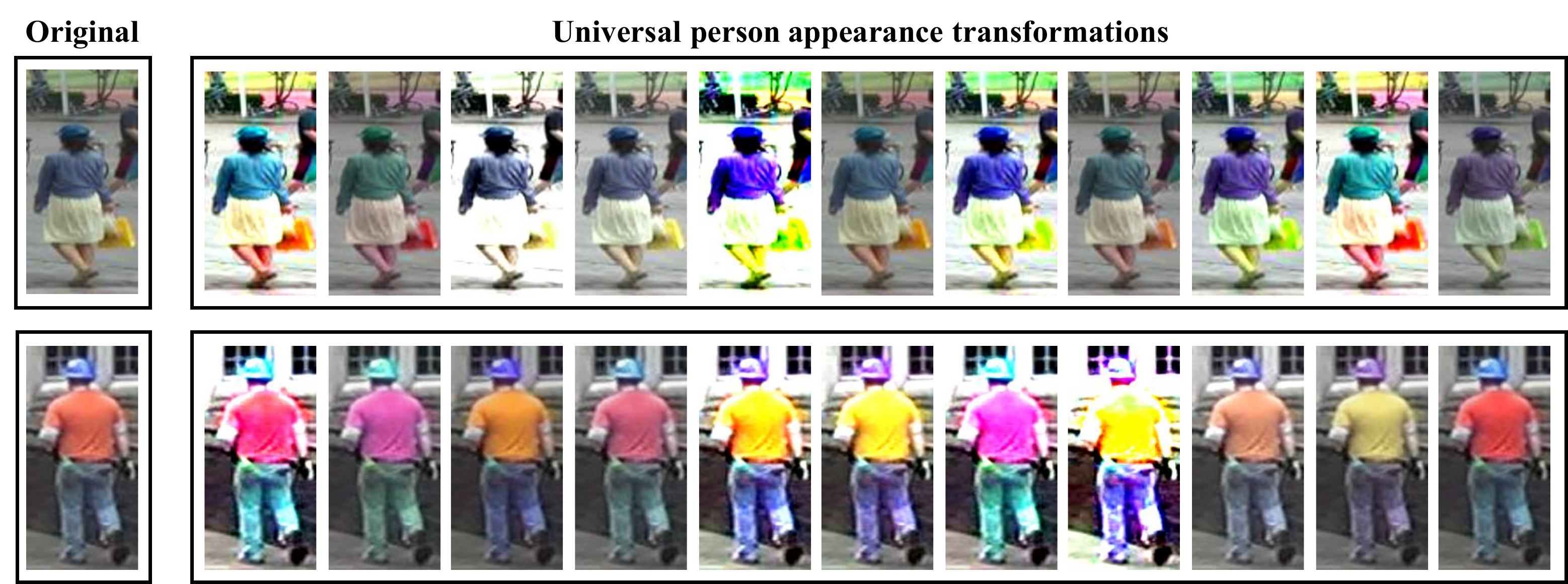}
		\captionof{figure}{Universal person appearance transformations
			for training a single domain-generic re-id model
			enabling universal deployments. Compared to existing
			methods typically focusing on domain-specific model training
			(``train once, run once''),
			the proposed method allows for a ``train once, run everywhere'' pattern
			therefore favourably  suits the industrial scale large system development
			without the need of training the system to every individual
			target domain as prior of each deployment.} \label{fig:teaser}
	\end{center}%
}]

\begin{abstract}
Most state-of-the-art person re-identification (re-id) methods
depend on supervised model learning 
with a large set of cross-view identity labelled training data.
Even worse, such trained models are limited to
only the {same-domain} deployment with significantly degraded
cross-domain generalisation capability, i.e. ``{\em domain specific}''.
To solve this limitation,
there are a number of recent unsupervised domain adaptation
and unsupervised learning methods that leverage
unlabelled target domain training data.
However, these methods need to train a separate model 
for each target domain as supervised learning methods. 
This conventional ``{\em train once, run once}'' pattern
is unscalable to a large number of target domains
typically encountered in real-world deployments.
We address this problem by
presenting a ``{\em train once, run everywhere}'' pattern
industry-scale systems are desperate for.
%
We formulate a ``{\em universal model learning}'' approach enabling domain-generic person re-id
using only limited training data of a ``{\em single}'' seed domain.
Specifically, we train a universal re-id deep model to 
discriminate between a set of transformed person identity classes.
Each of such classes is formed by applying a variety of random appearance transformations to the images of that class, where
the transformations simulate the camera viewing conditions of any domains
for making the model training domain generic.
%
%
%
%
Extensive evaluations show the superiority of our method 
for universal person re-id over a wide variety of state-of-the-art 
unsupervised domain adaptation
and unsupervised learning re-id methods on five standard benchmarks:
Market-1501, DukeMTMC, CUHK03, MSMT17, and VIPeR.
\end{abstract}

\section{Introduction}
\begin{figure} 
	\centering
	\includegraphics[width=1.0\linewidth]{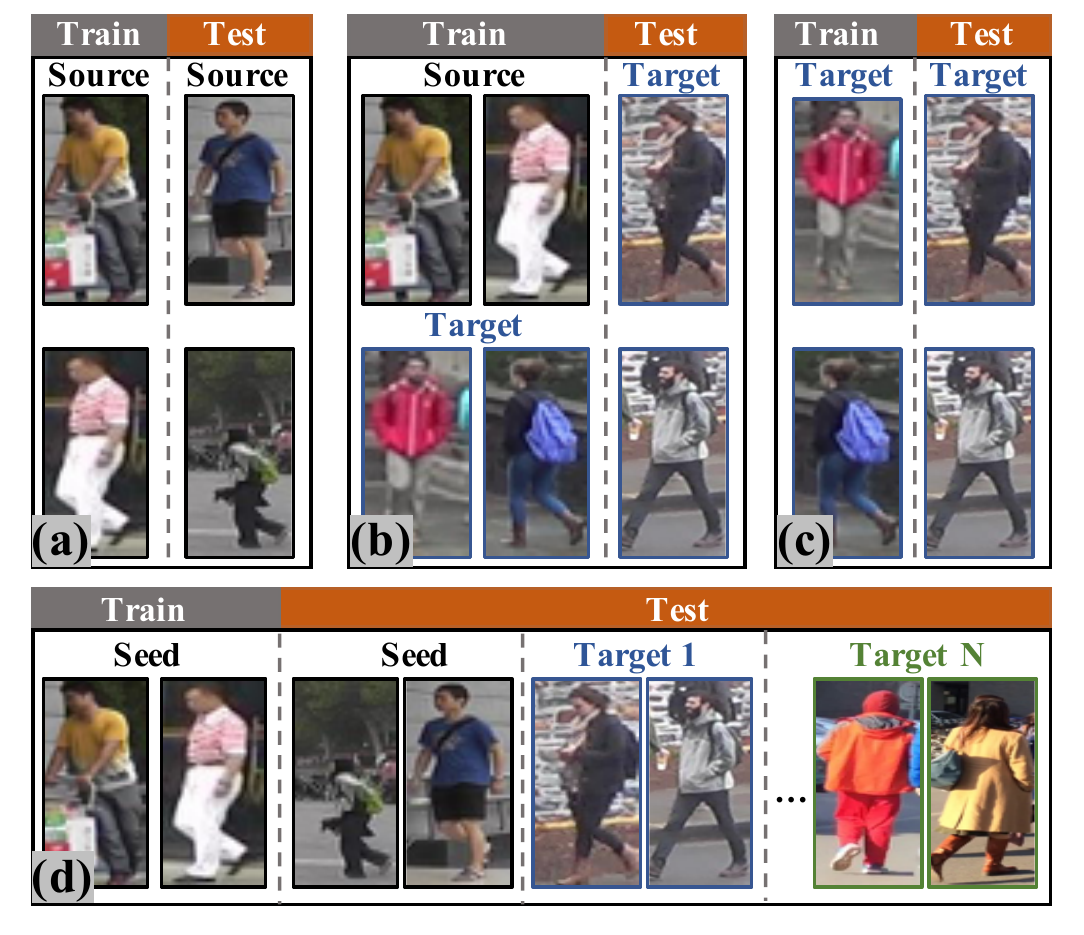} 	
	\caption{Four representative model learning strategies for person re-id:
		{\bf (a)} 
		{\em Supervised model learning} on a large set of
		cross-camera identity labelled training data per domain.
		Once trained, the model is deployed for the same domain alone.
		{\bf (b)} 
		{\em Unsupervised domain adaptation}
		on labelled training data from a source domain
		and unlabelled training data from the target domain.
		The adapted model is specific for the target domain.
		{\bf (c)} 
		{\em Unsupervised model learning} 
		on unlabelled training data of the target domain.
		The trained model is specific for the target domain.
		{\bf (d)}
		{\em Universal model learning} 
		on labelled training samples from a seed domain.
		Once trained the model can be {\em frozen forever}
		and applied for universal person re-id deployment at any target domains
		including the seed domain.
	}
	\label{fig:designs}
\end{figure}

\vspace{0.2cm}
The aim of person re-identification (re-id) is to recognise the 
fine-grained person identity information
in detected bounding boxes captured from
non-overlapping surveillance camera views
\cite{gong2014person}.
State-of-the-art re-id methods train deep Convolutional Neural Network (CNN) models in a {\em supervised learning} manner and dramatically improve the matching performance
\cite{sun2018beyond,li2018harmonious,zhang2017deep,li2017person,wei2017person,wang2018mancs,xu2018attention,song2018mask}.
%
Supervised model learning often assumes {\em only} the same-domain 
(i.e. surveillance camera network) deployment
with drastic performance drop for new unseen domains, 
as the model is trained specifically to well fit a large training set per domain
(Fig. \ref{fig:designs}(a)).
This restricts severely their scalability and deployability
in real-world applications where no manually labelled training data is typically available
for target domains.
Exhaustive cross-camera labelling 
is rather expensive and not always available, 
due to a quadratic number of camera view pairs in each
surveillance domain.

There are a small number of recent attempts that aim for solving 
the aforementioned generalisation limitations of supervised learning re-id methods.
Their high-level modelling strategies fall generally into three groups: 
(i) {\em hand-crafting features} 
\cite{farenzena2010person,cheng2011custom,liao2015person,zheng2015scalable},
(ii) {\em unsupervised domain adaptation}
\cite{peng2016unsupervised,zheng2017unlabeled,deng2018image,wang2018transferable,lin2018multi,zhong2018generalizing,yu2018unsupervised},
(iii) {\em unsupervised deep learning}
\cite{chen2018deep,lin2019aBottom,li2018unsupervised}. 
Hand-crafting feature representations are largely domain generic and universal
without the need for training. 
However, they suffer from much inferior re-id performances 
than the latter two approaches,
due to rather limited appearance knowledge involved.
Whilst significant performance gains have been achieved on unlabelled target domains,
existing unsupervised domain adaptation (Fig. \ref{fig:designs}(b)) and 
unsupervised model learning (Fig. \ref{fig:designs}(c)),
often take a ``{\em train once, run once}'' pattern.
That is, a trained model by them is effective only
for the target domain that the model training is applied to.
For every single target domain deployment, a new model needs to 
be trained through the same optimisation process repeatedly.
Such a {\em domain-specific} property reduces their practical value
and limits their scalability significantly, considering
potentially a very large quantity of different domains
to be targeted in real-world applications.


In this work, we consider a ``{\em train once, run everywhere}'' pattern.
In contrast to all the existing methods,
we train a re-id model using the labelled data from a single {\em seed} domain,
and {\em frozen} it for universal deployment at any domains
{\em without} further training and/or fine-tuning the model
to any target domains (Fig. \ref{fig:designs}(d)).
To this end, we propose a {\em Universal Model Learning} (UML) method
capable of training a {\em single} model
for domain generic person re-id deployment.
UML trains a universally deployable re-id model {\em one-off}
on transformed seed training data,
without the need of using any target domain data for model learning and refinement. 
The image transformations 
are designed to 
produce an extremely rich and diverse training dataset (Fig. \ref{fig:teaser}) that simulates 
camera viewing condition variations as completely as possible
for different domains, i.e. {\em domain complete therefore domain generic}.
The viewing condition variations are simulated by randomly applying 
fundamental colour and contrast transformations to a labelled 
seed person image.
This image and its transformed versions share the same identity label.
By design, the re-id model trained on 
the proposed augmented dataset 
is discriminative and effective universally for any domains.

Our {\em contributions} of this work are summarised as follows:
{\bf(1)} 
We present a ``train once, run everywhere''
pattern for universal person re-identification.
This differs dramatically from the conventional
state-of-the-art re-id methods in form of 
``train once, run once'', which is unsuitable and unscalable for
practical large scale system development and deployment
in industrial applications.
To our knowledge, this is the first deep learning attempt 
of universal person re-id.
{\bf(2)} 
We propose a simple yet effective {\em Universal Model Learning}
(UML) approach for realising universal person re-id.
UML is based on comprehensive and rich image transformations
in colour and contrast given a labelled seed training dataset.
This image generation method is computationally efficient,
flexible in design, and domain generic, without the need for complicated image synthesis
model design and intricate hyper-parameter tuning per domain
as required by state-of-the-art methods.
{\bf(3)} 
We summarise the existing methods of
unsupervised domain adaptation and unsupervised learning
applicable for person re-id in unlabelled target domains,
and compare them with the proposed UML method
in proper perspective.
Extensive evaluations demonstrate the model training and performance superiority 
of our method over the state-of-the-art alternative methods on five person re-id benchmarks: 
Market-1501, DukeMTMC-reID, CUHK03, MSMT17,
and VIPeR.

\section{Related Work}
\vspace{0.4cm}
\noindent {\bf Supervised person re-id.}
In the literature, existing person re-id methods
mostly focus on {\em supervised model learning}
\cite{gong2014person,li2014deepreid,zheng2015scalable,li2017person,chen2018deep,chen2017personICCVWS,wang2014person,xiao2016learning,wang2018mancs,li2018harmonious,chang2018MLFN_reid,sun2018beyond,chen2017person,Yu_2018_ECCV,Shen_2018_ECCV,xu2018attention,wang2018learning}.
This type of methods requires
to access a large set of cross-camera identity labelled
training samples acquired through an exhaustive and costly annotation process.
Deploying a pre-trained re-id model to unseen 
new domains often encounter dramatic performance drops.
Due to this prerequisite for model supervised optimisation,
their scalability and usability are dramatically restricted
in real-world applications.
This is because, we often have no access to 
any cross-camera identity labelled training data
for target deployment domains in practical use.
Earlier hand-crafted features based models \cite{farenzena2010person,cheng2011custom,liao2015person,zheng2015scalable,zhao2017person}
are domain generic,
but yield significantly inferior model performance.

\vspace{0.2cm}
\noindent {\bf Unsupervised domain adaptation person re-id.}
The limitation of supervised learning re-id methods
in cross-domain scalability can be addressed 
by using unsupervised domain adaptation (UDA) techniques.
Existing UDA re-id methods generally fall into two categories:
(1) image synthesis \cite{deng2018image,zhong2018generalizing,bak2018domain,qian2018pose}, and 
(2) feature alignment \cite{peng2016unsupervised,yu2018unsupervised,yu2017cross,wang2018transferable,lin2018multi}.
The former aims to transfer the labelled source identity classes
from the source domain to the target domain
through cross-domain conditional image generation
in the appearance style and background context at pixel level.
The synthetic images are then used to fine-tune
the model towards the target domain.
On the contrary, the latter
transfers the discriminative feature information
learned from the labelled source training data
to the target feature space by distribution
alignment.
These methods often use discrete attribute labels
for facilitating the information transition 
across domains due to their better domain invariance
property than low-level feature representations.

\vspace{0.2cm}
\noindent {\bf Unsupervised learning person re-id.} 
In parallel to UDA re-id methods,
unsupervised learning methods have started to gain
increasing potentials for eliminating the need of labelling 
large training data \cite{kodirov2015dictionary,lisanti2015person,li2018unsupervised,lin2019aBottom,chen2018deep}.
Such methods rely on the reconstruction loss designs
\cite{kodirov2015dictionary,lisanti2015person}
or self-discovered cross camera label information
by the in-training model for self-supervised learning
\cite{li2018unsupervised,lin2019aBottom,chen2018deep}.
The methods assume access of unlabelled training data
sampled from the target domain, and train
a domain-specific re-id model.

\vspace{0.35cm}
\noindent {\bf Universal learning person re-id.} 
In this work, we present universal learning person re-id,
which differs dramatically from all the existing methods
as discussed above.
Our method trains a single {\em domain-generic}
re-id model for universal deployments.
This is in contrast to previous learning algorithms
usually producing {\em domain-specific} models
using either labelled source and/or unlabelled target domain training data.
That being said, a separate model training is required 
for each target domain deployment 
which is neither cost-effective and convenient nor
allowed always for industrial settings. 
The re-id model trained by our method
can be immediately deployed to any domains
where no any video and image data are observed 
to model optimisation.
%
Such universal deployment property is favourable
and desired to practical system development.
Moreover, the proposed image transformation method is 
computationally efficient with flexible design
due to no need for complex model formulation
and costly pixel synthesis model training.
In comparison to hand-crafted features \cite{gray2008viewpoint,farenzena2010person,liao2015person,cheng2011custom,zheng2015scalable}, 
our model has extra capability 
for feature representation learning and model optimisation
as supervised and unsupervised learning counterparts,
whilst simultaneously retaining the merits 
of domain universality as hand-crafted features.
Besides, our learning method
differs from
and is more scalable than
multi-target domain adaptation \cite{yu2018multi,gholami2018unsupervised}
where all the target domains need 
to be observed to training.
Conceptually, our method generalises 
the notion of multi-target domain simultaneous adaptation since we make a model effective for all different domains even without accessing any target domain data.

\section{Methodology}
\vspace{0.4cm}
\begin{figure} 
	\centering
	\includegraphics[width=0.85\linewidth,height=0.7\linewidth]{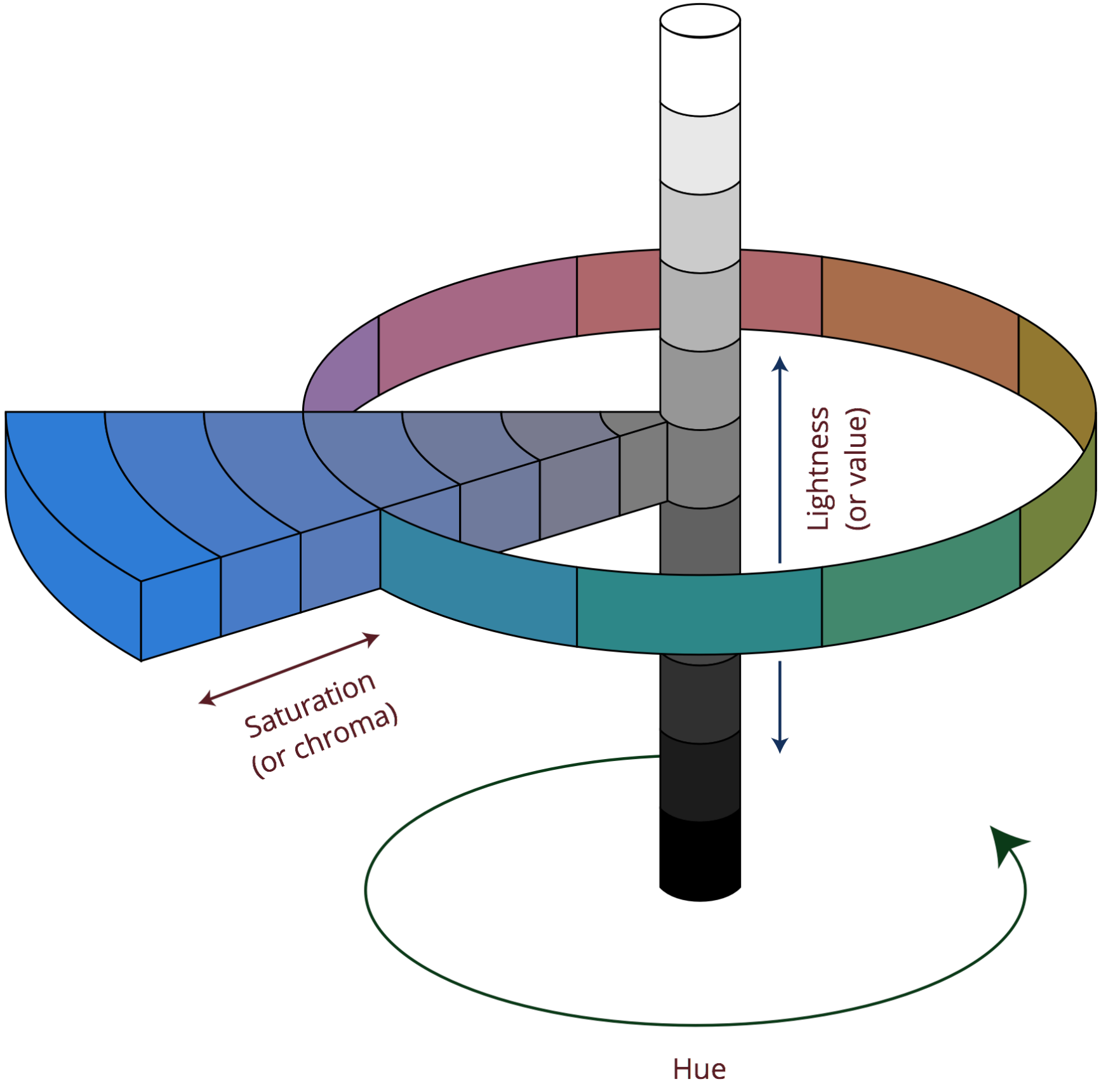} 	
	\caption{
		Illustration of Munsell colour system in three dimensions: Hue, Saturation, Lightness.
		This graph is adopted from \cite{Munsell}.
		Best viewed in colour.
	}
	\label{fig:color}
		\vspace{-0.3cm}
\end{figure}

\begin{figure*} 
	\centering
	\includegraphics[width=1.0\linewidth]{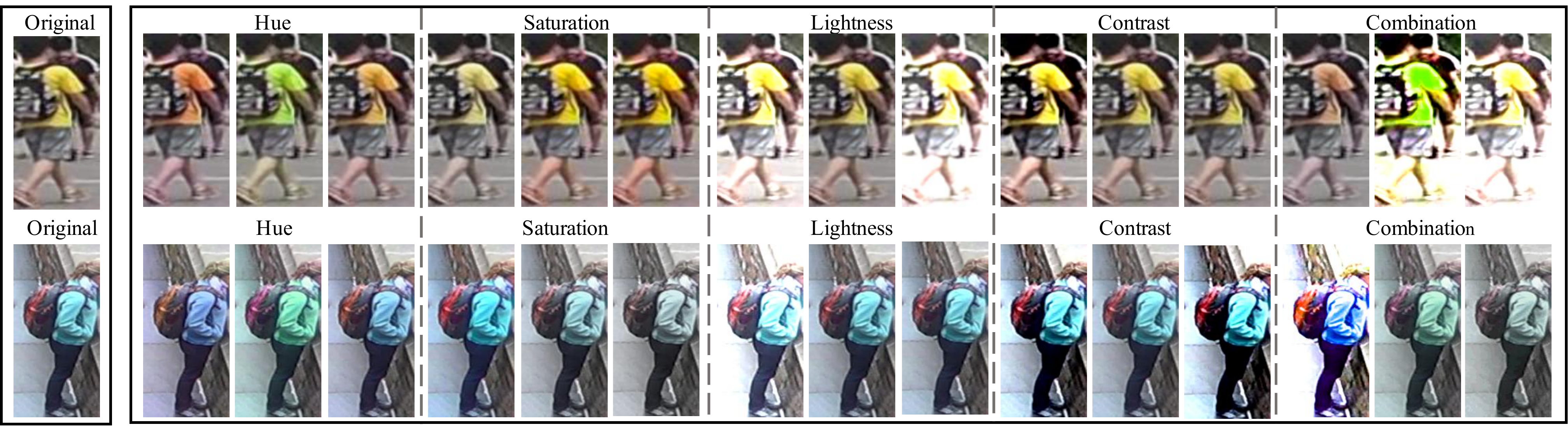} 	
	\caption{
		Example transformations in Hue, Saturation (Chroma), Lightness (Value), 
		Contrast, and their random combinations.
	}
	\label{fig:ILR}
\end{figure*}

\subsection{Universal Image Transformation}
\vspace{0.3cm}
To train a universal person re-id network model,
we assume a labelled seed training dataset (i.e. a seed domain)
$\mathcal{I}=\{\bm{I}_i,y_i\}_{i=1}^{N}$,
consisting of $N$ person 
bounding box images $\bm{I}_i$ each
annotated with 
a person {identity} class label $y_i \in \{1,\cdots, N_\text{id}\}$,
It contains a total of $N_\text{id}$ different person identities.

We propose to transform this seed training set $D$
so as to cover the camera viewing conditions of 
arbitrary domains.
Formally, we define a set of transformations
$\{\mathcal{M}_{\bm{t}}\}, \bm{t} \in \mathcal{T}$
where $\bm{t}$ defines the transformation parameter vector
and $\mathcal{T}$ the transformation space.
Each transformation $\mathcal{M}_{\bm{t}}$ 
is composited of several primitive transformations.
Consider the variations of person appearance at typical surveillance scenes
are largely due to illumination (lighting), 
we establish a space of linear transformations
with regard to pixel colour and contrast.
We note that the approach is flexible 
to adopt other transformations in order to better serve other target domains.

Specifically, we consider the colour transformations
in the HSV representation space \cite{munsell1919color}.
Each colour has three fundamental attributes (Fig. \ref{fig:color}):
\begin{enumerate}
	\item {\em Hue}: Colour such as red, orange, yellow, and so forth.
	It depends on the wavelength of light reflected and/or produced.
	
	\item {\em Saturation} ({\em Chroma}): The brilliance of a hue,
	i.e. how pure (intense) a hue is. 
	More saturated a hue is, 
	brighter it appears.
	
	\item {\em Lightness} ({\em Value}): The lightness or darkness of a hue.
	Adding white (black) makes the colour lighter (darker).
	Note, the effect of lightness is relative to other values in a composition.
\end{enumerate}
%


Specifically, 
for hue transformation we convert 
the image into HSV and
add the corresponding parameter to the original value on the hue dimension.
Afterwards, the image is converted back. 
For the other factors (including contrast), we
perform the transformation
by linear interpolation and extrapolation \cite{haeberli1994image}. For restricting the transformations to perceptually sensible scope,
we define the variation range as: 
hue in [-18, 18] (cyclical), 
saturation in [0.6, 1.4], 
lightness in [0.6, 1.4], and 
contrast [0.6, 1.4].
For saturation/lightness/contrast, the value of ``1'' means {\em no} transformation, and 
for hue ``0'' means {\em no} transformation.

To form a single colour-contrast transformation, we 
sample a parameter value for each factor 
and concatenate them into a vector $\bm{t}$.
We consider an online image transformation strategy for stochastic deep learning.
This avoids the need for saving and managing a large quantity
of transformed images.
In a training iteration, given a seed image $\bm{I}_i$,
we sample a randomly parameter vector
$\bm{t}_i$
and apply the corresponding transformation
$\mathcal{M}_{\bm{t}_i}$ to it.
As such, we obtain a 
transformed variant:
\begin{equation}
D_i = \bm{I}_i\mathcal{M}_{\bm{t}_i}.
\end{equation}

By repeating the transformation on each and every person image 
of a training mini-batch,
we form domain generic universal 
training samples $\{D_i\}$ for model training on-the-fly.
%
We show examples of transformed person images in
Fig \ref{fig:ILR}.
Perceptually, such transformations leave
the original identity class information of person images intact, facilitating the re-id discriminative model optimisation.

\subsection{Person Re-Identification Model}
\vspace{0.3cm}
For person re-id model,
we use ResNet-50 \cite{he2016deep} as the backbone network.
Other networks \cite{chang2018MLFN_reid,li2018harmonious,zhang2017deep}
can be readily considered without any restriction.
To enable fine-grained part-level discriminative learning, 
we adopt the PCB design \cite{sun2018beyond}. 
Instead of the whole image, 
PCB uses average pooling on local regions and 
applies a separate re-id loss supervision 
on each individual region independently and concurrently.
In addition, we add a parallel global branch 
for discriminative learning of the whole images.
We apply label smoothing for mitigating
model overfitting \cite{szegedy2016rethinking}.
%
For model training, we adopt the softmax Cross Entropy loss as the objective function: 
\begin{equation}
\mathcal{L}_\text{ce} = - 
\sum_{k=1}^{N_\text{id}}\sum_{j=1}^{m} \delta_{k,y}  \log \big(p_{j}({k}|\bm{I}_{i}) \big)
+ \log \big(p({k}|\bm{I}_{i}) \big)
\label{eq:cross_loss}
\end{equation}
where  $\delta_{k,y}$ is the Dirac delta returning $1$ 
if $k$ is the ground-truth class label $y$, otherwise 0.
$p_j$ and $p$ denotes the class posterior probability
of the $j$-th local region and the whole image,
estimated by the current network. 
$m$ indicates the total number of local regions.
We set $m=6$ in our experiments the same as \cite{sun2018beyond}.

In test, we concatenate all the local regional and global 
features as the final re-id representation.
We adopt the Euclidean distance metric for re-id matching and ranking.

\subsection{Remarks}
\vspace{0.3cm}
Compared to previous data augmentation approaches 
\cite{krizhevsky2012imagenet,russakovsky2015imagenet}, 
our method differently focuses on training a 
universal model for any domain generalisation,
other than enriching domain-specific training data variety 
and learning a better model for that domain alone.
In particular, we uniquely consider training data transformations
that simulate the person appearance distributions and characteristics of 
arbitrary unseen domains.
Such data augmentation is not necessarily beneficial
for the labelled training data domain.

As hand-crafted feature representations \cite{gray2008viewpoint,farenzena2010person,cheng2011custom,liao2015person,zheng2015scalable},
our model learning is universal and domain generic with 
favourable cross-domain scalability and usability.
However, our model is much more discriminative
due to the deep model learning capability 
that is able to learn powerful feature representations.
On the contrary, hand-crafted features are based on limited
human knowledge which offers significantly weaker 
discrimination performance.

The closest works to our method 
are image synthesis based unsupervised domain adaptation re-id models 
\cite{bak2018domain,qian2018pose,zhong2018generalizing,zheng2017unlabeled,deng2018image}.
All of them aim to transfer the labelled source person identity information
to unlabelled target domains.
However, these existing methods are domain-specific,
and often need a complex model training for every single target domain.
This ``train once, run once'' strategy is less usable and more costly to 
real-world system development.
In contrast, our method needs neither domain-specific training
nor difficult model optimisation (such as GANs \cite{goodfellow2014generative}).
We take a ``train once, run everywhere'' strategy
based on simple and domain universal image transformations.
Our method uses flexibly off-the-shelf supervised learning re-id methods 
therefore can benefit continuously from
a wide range of increasingly advanced learning algorithms
developed by the wider community.

\begin{table} 
	\centering
	\setlength{\tabcolsep}{0.02cm}
	\begin{tabular}{l||c|c|c|c}
		\hline 
		\multirow{2}{*}{Dataset}
		& \multicolumn{2}{c|}{Train}  & \multicolumn{2}{c}{Test}\\
		\cline{2-5}
		& \# ID  & \# Image	& \# ID & \# Image \\
		\cline{1-5}
		VIPeR \cite {gray2008viewpoint}
		& 316 & 632 & 316 & 632 \\
		CUHK03 \cite{li2014deepreid}
		& 767 & 7,368 & 700 & 6,728 \\
		Market-1501 \cite{zheng2015scalable}
		& 751 &12,936  & 750 &19,732 \\
		DukeMTMC \cite{zheng2017unlabeled}
		& 702 &16,522 & 702 &18,889 \\
		MSMT17 \cite{wei2018person}
		& 1,041 & 32,621 & 3,060 & 93,820
		\\ \hline
	\end{tabular}
	\vspace{0.2cm}
	\caption{
		Dataset statistics and evaluation setting.
	}
	\label{tab:dataset_stats}
\end{table}
\vspace{-0.2cm}

\begin{figure} [h]
	\centering
	\includegraphics[width=1.0\linewidth]{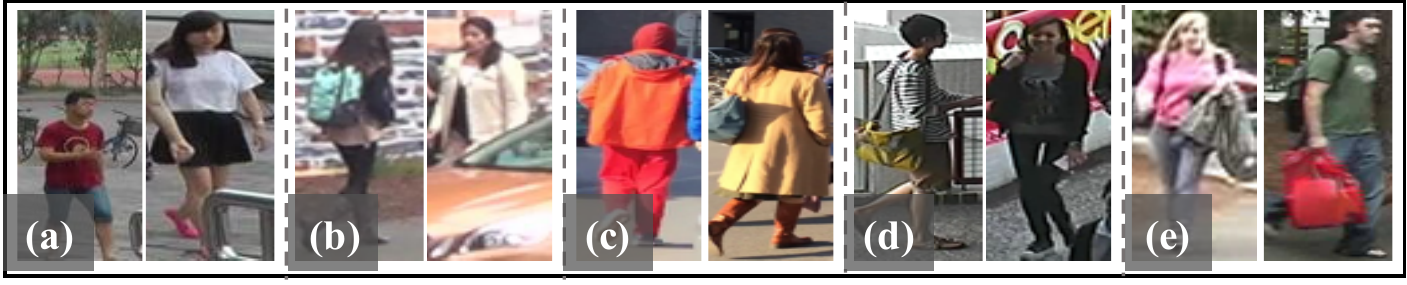}
	\caption{Image examples from the
		(a) Market-1501, (b) DukeMTMC-reID, 
		(c) MSMT17,
		(d) CUHK03, and (e) VIPeR
		person re-id datasets.}
	\label{fig:dataset}
	\vspace{-0.2cm}
\end{figure}

\section{Experiment}
\vspace{0.4cm}
\noindent{\bf Datasets.}
To evaluate our model, we tested five popular person re-id benchmarks.
We used the standard train/test evaluation
protocols.
The statistics of these datasets are summarised 
in Table \ref{tab:dataset_stats}.
Example person images of these datasets
are shown in Fig \ref{fig:dataset}.


\vspace{0.2cm}
\noindent{\bf Performance metrics.}
We adopted the Cumulative Matching Characteristic (CMC) and mean Average Precision (mAP) 
as the model performance measurements.

\vspace{0.2cm}
\noindent{\bf Implementation details.}
We conducted all the experiments in PyTorch \cite{paszke2017automatic}. 
To train a universal re-id model,
we initialised the model with the
parameters pre-trained on ImageNet and used the SGD algorithm with the momentum set to $0.9$, the weight decay to 0.0005, 
and the mini-batch size of 32.
We trained the model for totally 60 epochs, with the learning rate of 0.001 
in the first 40 epochs, and the decay learning rate as 10 in the last 20 epochs. 
All input images were resized to 384$\times$128 in pixel 
and subtracted by the ImageNet mean. 
On top of the proposed transformation strategy, 
we applied random cropping and flipping during training.


\begin{table*} 
	\setlength{\tabcolsep}{0.05cm} 
	\caption{
		Universal learning {\em vs.} unsupervised learning.
	}
	\vskip 0.1cm
	\label{tab:unsupervised-reid}
	\centering
	\begin{tabular}{l|ccc|c||ccc|c||cccc||ccc|c ||ccc}
		\hline 
		Method
		&\multicolumn{4}{c||}{Market-1501}
		&\multicolumn{4}{c||}{Duke}
		&\multicolumn{4}{c||}{MSMT17}
		&\multicolumn{4}{c||}{CUHK03} 
		&\multicolumn{3}{c}{VIPeR}\\ 
		\hline
		Metric(\%) 
		&R1 & R5 & R10 & mAP 
		&R1 & R5 & R10 & mAP
		&R1 & R5 & R10 & mAP 
		&R1 & R5 & R10 & mAP
		&R1 & R5 & R10 \\
		\hline
		\hline 
		LOMO \cite{liao2015person} 
		&27.2&41.6&49.1&8.0
		&12.3&21.3&26.6&4.8
		&-&-&-&-
		&0.6&1.9&3.6&0.7&-&-&-\\
		BOW \cite{zheng2015scalable}
		& 35.8 & 52.4 & 60.3 & 14.8
		& 17.1 & 28.8 & 34.9 & 8.3
		&-&-&-&-
		&2.1 & 4.6 & 7.0 & 1.9&-&-&-\\
		\hline    
		ISR \cite{lisanti2015person}
		&40.3 &-&-&14.3&-
		&-&-&-&-
		&-&-&-&-
		&-&-&-&27.0&49.8&61.2\\
		Dic \cite{kodirov2015dictionary}
		& 50.2 & - & - & 22.7 &-&-&-&-&-&-&-&-&-&-&-&-&29.6&54.8&64.8\\
		\hline
		TAUDL \cite{li2018unsupervised}
		&63.7&-&-&\bf41.2
		&\bf61.7&-&-&\bf43.5
		&28.4&-&-&12.5
		&-&-&-&-&-&-&-\\
		BUC \cite{lin2019aBottom}
		&66.2 &79.6 &84.5 & 38.3
		& 47.4 & 62.6 & 68.4 & 27.5
		&-&-&-&-
		&-&-&-&-&-&-&- \\
		\hline 
		\bf UML(MSMT17)&\bf68.2&\bf83.1&\bf87.6&37.0&60.9&\bf83.0&\bf87.5&37.0&-&-&-&-&\bf12.6&\bf25.4&\bf33.4&\bf13.4&\bf36.4&\bf57.9&\bf67.4
		\\
		
		\bf UML(Duke)&
		66.1&81.6&86.3&35.5
		&-&-&-&-
		&\bf35.5&\bf48.2&\bf53.7&\bf12.2
		&10.7&21.9&27.5&10.5&35.4&54.1&62.3
		\\
		\hline \hline
		\em Supervised Learning
		&90.4 &96.5&97.8&73.5	& 81.5&  90.8&  93.1&65.9
		&73.3 &84.8&88.1&44.1&40.8&62.7&73.6&40.4&39.2&65.8&77.5
		\\ \hline
	\end{tabular}
\end{table*}

\subsection{Universal Learning {\em vs.} Unsupervised Learning}
\vspace{0.2cm}
We compared our UML model with 
two hand-crafted feature models, 
(LOMO \cite{liao2015person}, 
BoW \cite{zheng2015scalable}),  
two dictionary learning models
(ISR \cite{lisanti2015person},
Dic \cite{kodirov2015dictionary}),
and 
two unsupervised deep learning methods
(TAUDL \cite{li2018unsupervised},
BUC \cite{lin2019aBottom}).
In this test, we used MSMT17 and DukeMTMC-reID as 
the seed training data, individually.
Table \ref{tab:unsupervised-reid} compares the performance of these methods.
We have the following findings.

\vspace{0.15cm}
{\bf(1)} Hand-crafted feature based re-id methods \cite{liao2015person,zheng2015scalable} 
produce the weakest matching performance.
This is due to poor representation capability of 
non-leaning based features without the ability
to extract data relevant feature patterns.
Also, these methods cannot optimise the matching metrics.

\vspace{0.15cm}
{\bf(2)}
Dictionary learning based methods 
\cite{lisanti2015person,kodirov2015dictionary}
improve the performance by using reconstruction
based learning objective loss functions. However, their capability 
is limited by the input hand-crafted feature representations.

\vspace{0.15cm}
{\bf(3)}
The more recent unsupervised deep learning models
\cite{li2018unsupervised,lin2019aBottom}
further push the performance envelope. 
In addition to per-domain model training requirement,
these methods often come with some extra model parameters
which are likely to be data sensitive.
Typically, careful parameter tuning and costly model training 
are required in order to 
achieve competitive results.
This is not favourable particularly
for unsupervised learning where {\em no}
labelled validation data available for hyper-parameter
cross-validation and optimisation
for the target domain.

\vspace{0.15cm}
{\bf(4)}
The proposed UML method matches or surpasses
the performance of best competitors \cite{li2018unsupervised,lin2019aBottom}
{\em without} training the model to the target domains.
This suggests stronger domain generalisation 
and practical advantages of our method for the industrial adoption
due to the ``train once, run everywhere'' merit.
In reality, model training is costly in both budget and time.
This therefore suggests an economical advantages and
deployment-friendly superiority
of our method over the strongest competitors in practice.

\vspace{0.15cm}
{\bf(5)} Using MSMT17 as the seed training data
leads to slightly better performance as
compared to selecting DukeMTMC. 
This is reasonable since MSMT17 offers more person identities and images. 

\vspace{0.15cm} 
{\bf(6)}
Compared to supervised learning,
both unsupervised and universal learning
models are significantly outperformed.
This indicates a large room for further algorithm innovation.

\begin{table*}[h]	
	\setlength{\tabcolsep}{0.06cm} 
	\caption{
		Universal learning {\em vs.} unsupervised domain adaptation.
		*: Using more labelled source training data.
		$^{\dagger}$: Using additionally person attribute labels.
		UML uses the source data as the seed training samples
		for fair comparison.
	}
	\label{tab:UDA}
	\centering
	\begin{tabular}{l|ccc|c||ccc|c||ccc|c||ccc|c}
		\hline 
		Source$\rightarrow$Target  
		&\multicolumn{4}{c||}{Duke$\rightarrow$Market}
		&\multicolumn{4}{c||}{Market$\rightarrow$Duke}
		&\multicolumn{4}{c||}{CUHK03$\rightarrow$Market}
		&\multicolumn{4}{c}{CUHK03$\rightarrow$Duke} \\ 
		\hline
		Metric (\%) 
		&R1 & R5 & R10 & mAP 
		&R1 & R5 & R10 & mAP
		&R1 & R5 & R10 & mAP 
		&R1 & R5 & R10 & mAP  \\
		\hline
		\hline
		UMDL$^{\dagger}$ \cite{peng2016unsupervised}
		&34.5&52.6&59.6&12.4&18.5&31.4&37.6&7.3
		&-&-&-&-
		&-&-&-&-\\
		PUL \cite{fan2017unsupervised}
		&45.5&60.7&66.7&20.5&30.0&43.4&48.5&16.4
		&41.9&57.3&64.3&18.0&23.0&34.0&39.5&12.0\\
		CAMEL* \cite{yu2017cross}
		&54.5&-&-&26.3
		&-&-&-&-
		&-&-&-&-
		&-&-&-&-\\
		TJ-AIDL$^{\dagger}$ \cite{wang2018transferable}
		&58.2&74.8&81.1&26.5&44.3&59.6&65.0&23.0
		&-&-&-&-&-&-&-&-\\
		MMFA$^{\dagger}$ \cite{lin2018multi}
		&56.7&75.0&81.8&27.4
		&45.3&59.8&66.3&24.7
		&-&-&-&-&-&-&-&-	\\
		DECAMEL* \cite{yu2018unsupervised}
		& 60.2 & - & - & 32.4
		& - & -&-&-& - & -&-&-& - & -&-\\
		\hline
		PTGAN \cite{wei2017person}
		&38.6&-&66.1&-&27.4&-&50.7&-
		&31.5&-&60.2&-&17.6&-&38.5&-\\
		PoseNorm \cite{qian2018pose}
		& - &- &- &-
		& 29.9 &- & 51.6 & 15.8 
		& - &- &- &-
		& - &- &- &- \\
		SPGAN \cite{deng2018image}
		&51.5&70.1&76.8&22.8
		&41.1&56.6&63.0&22.3
		&42.3&-&-&19.0&-&-&-&-\\
		SyRI* \cite{bak2018domain}
		& 65.7 &- &- &-
		& - &- &- &-
		& - &- &- &-
		& - &- &- &- \\
		HHL \cite{zhong2018generalizing}
		&62.2&78.8 &84.0 &31.4
		&\bf46.9 &61.0 &66.7 &\bf27.2
		&56.8&74.7&81.4&29.8&42.7&57.5&64.2&23.1 \\
		\hline
		\bf UML(Source) 
		&\bf66.1&\bf81.6&\bf86.3&\bf35.5 &46.3&\bf61.2&\bf66.8&26.7&\bf58.7&\bf76.5&\bf82.6&\bf31.1&\bf42.8&\bf57.8&\bf64.3&\bf23.2\\
		\hline \hline
		\em Supervised Learning
		& 90.4&96.5&97.8&73.5
		&81.5 & 90.8 & 93.1 & 65.9
		& 90.4&96.5&97.8&73.5
		&81.5 & 90.8 & 93.1 & 65.9 
		\\ \hline
	\end{tabular}
	\vspace{-0.2cm}
\end{table*}

\subsection{Universal Learning {\em vs.} Domain Adaptation}\vspace{0.2cm}
We compared our UML with the state-of-the-art unsupervised
domain adaptation re-id methods, including
five image synthesis models
(PTGAN \cite{wei2017person}, 
PoseNorm \cite{qian2018pose},
SyRI \cite{bak2018domain},
SPGAN \cite{deng2018image},
HHL \cite{zhong2018generalizing}),
and 
six feature alignment models
(UMDL \cite{peng2016unsupervised}
CAMEL \cite{yu2017cross},  
PUL \cite{fan2017unsupervised},
TJ-AIDL \cite{wang2018transferable}, 
MMFA \cite{lin2018multi},
DECAMEL \cite{yu2018unsupervised}).
We make several observations from Table \ref{tab:UDA} as below. 

\vspace{0.15cm}
{\bf(1)}
Feature alignment methods have obtained
increasingly higher performance.
It is worth noting that most feature learning methods such as DECAMEL unfairly benefit from extra labelled source data.

\vspace{0.15cm}
{\bf(2)}
Compared to feature alignment,
image synthesis methods have started to
achieve relatively superior cross-domain 
re-id accuracy.
It is especially so considering
that less label supervision is used
(except SyRI).

\vspace{0.15cm}
{\bf(3)}
UML is the best performer overall 
among all competitors. 
Importantly, unlike previous approaches for 
{\em domain-specific} model training,
our method needs only one time of 
{\em domain-generic} model training.
This uniquely enables 
universal person re-id deployments.

\vspace{0.15cm}
{\bf(4)}
Both unsupervised domain adaptation and universal learning
are significantly outperformed by 
the less scalable supervised learning,
suggesting the necessity of devoting further more
research efforts and endeavour for scaling
state-of-the-art re-id methods.

\subsection{Further Analysis and Discussions}

\vspace{0.2cm}
\noindent{\bf Effect of universal image transformations.} We evaluated the performance effect of the
proposed universal image transformations.
To this end, we compared the results
without using our transformations
on training data.
We tested Market-1501 and DukeMTMC-reID as
the seed domain, respectively.
Table \ref{tab:UML} shows that
the proposed image transformation
is consistently beneficial for improving 
the model performance on diverse 
target domains with very different camera viewing
conditions.
Both the absolute and relative performance gains
are significant in most cases.
As we aim for a domain-generic universal 
person re-id, the performance may be
inferior to that of domain-specific models
such as supervised learning models.
To examine this, we compared UML with the
supervised learning model (see the part with grey background).
We indeed observe a performance drop
but importantly not significant. 
This means that our model can be similarly
effective for the seed domain 
as the supervised learning method.
This differs from most image synthesis adaptation
methods which often suffer dramatic
performance decrease on the source domain.
For example, SPGAN \cite{deng2018image} experiences a Rank-1 drop of 17.2\% (91.6\%-74.4\%) and 16.4\% (83.5\%-67.1\%)
on the source Market-1501 and DukeMTMC-reID,
separately.

\begin{table*}[h]	
	\setlength{\tabcolsep}{0.015cm} 
	\caption{
		Effect of universal image transformations (UIT).
	}
	\vskip 0.1cm
	\label{tab:UML}
	\centering
	\begin{tabular}{l|cccc||cccc||cccc||cccc||ccc}
		\hline 
		\multirow{2}{*}{Seed}
		&\multicolumn{4}{c||}{Market}
		&\multicolumn{4}{c||}{Duke}
		&\multicolumn{4}{c||}{MSMT17}
		&\multicolumn{4}{c||}{CUHK03}
		&\multicolumn{3}{c}{VIPeR} \\ 
		\cline{2-20}
		&R1 & R5 & R10 & mAP 
		&R1 & R5 & R10 & mAP
		&R1 & R5 & R10 & mAP 
		&R1 & R5 & R10 & mAP
		&R1 & R5 & R10 
		\\
		\hline \hline
		
		Market
		&\bf\cellcolor{lightgray}91.6&\bf\cellcolor{lightgray}97.1&\bf\cellcolor{lightgray}98.0&\bf\cellcolor{lightgray}75.9
		&40.1&54.9&61.2&23.4
		&16.0&24.8&29.4&5.0
		&7.6&15.6&21.1&7.5&29.4&43.4&52.5
		\\
		\bf Market+UIT
		&\cellcolor{lightgray}90.4&\cellcolor{lightgray}96.5&\cellcolor{lightgray}97.8&\cellcolor{lightgray}73.5
		&\bf46.3&\bf61.2&\bf66.8&\bf26.7
		&\bf24.4&\bf35.7&\bf41.0&\bf7.6
		&\bf10.0&\bf20.1&\bf26.0&\bf9.9&\bf33.2&\bf48.1&\bf58.2
		\\
		\hline
		Gain(absolute)
		&\cellcolor{lightgray}-1.2&\cellcolor{lightgray}-0.6&\cellcolor{lightgray}-0.2&\cellcolor{lightgray}-2.4&+6.2&+6.3&+5.6&+3.3&+8.4&+10.9&+11.6&+2.6&+2.4&+4.5&+4.9&+2.4&+3.8&+4.7&+5.7
		\\
		Gain(relative)
		&\cellcolor{lightgray}-1.3&\cellcolor{lightgray}-0.6&\cellcolor{lightgray}-0.2&\cellcolor{lightgray}-3.2&+15.5&+11.5&+9.2&+14.1&+52.5&+44.0&+39.5&+52.0&+31.6&28.9&+23.2&+32&+12.9&+10.8&+10.9
		\\
		\hline \hline
		Duke
		&57.0&73.9&80.2&30.1
		&\bf83.5\cellcolor{lightgray}&\bf91.8\cellcolor{lightgray}&\bf94.3\cellcolor{lightgray}&\bf70.0\cellcolor{lightgray}
		&24.2&34.9&40.3&8.1
		&8.4&18.2&25.1&8.6&28.2&46.8&57.6
		\\
		\bf Duke+UIT
		
		&\bf66.1&\bf81.6&\bf86.3&\bf35.5
		&81.5\cellcolor{lightgray}&90.8\cellcolor{lightgray}&93.1\cellcolor{lightgray}&\cellcolor{lightgray}65.9
		&\bf35.5&\bf48.2&\bf53.7&\bf12.2
		&\bf10.7&\bf21.9&\bf27.5&\bf10.5&\bf35.4&\bf54.1&\bf62.3
		\\
		\hline 
		Gain(absolute)
		&+9.1&+7.7&+6.1&+5.4&\cellcolor{lightgray}-2.0&\cellcolor{lightgray}-1.0&\cellcolor{lightgray}-1.2&\cellcolor{lightgray}-4.1&+11.3&+13.3&+13.4&+4.1&+2.3&+3.7&+2.4&+1.9&+7.2&+7.3&+4.7
		\\
		\hline 
		
		Gain(relative)
		&+16.0&+10.4&+7.6&+17.9&\cellcolor{lightgray}-2.4&\cellcolor{lightgray}-1.1&\cellcolor{lightgray}-1.3&\cellcolor{lightgray}-5.9&+46.7&+38.1&+33.3&+50.6&+27.4&+20.3&+9.6&+22.1&+25.5&+15.6&+8.2
		\\
		\hline 	
	\end{tabular}
	
\end{table*}

\begin{table*}[h]	
	\setlength{\tabcolsep}{0.09cm} 
	\caption{
		Effect of individual image transformations: 
		{\underline H}ue, {\underline S}aturation, 
		{\underline L}ightness, and {\underline C}ontrast.
	}
	\vskip 0.1cm
	\label{tab:each_transf}
	\centering
	\begin{tabular}{l|cccc||cccc||cccc||cccc||ccc}
		\hline 
		\multirow{2}{*}{Seed}
		&\multicolumn{4}{c||}{Market}
		&\multicolumn{4}{c||}{Duke}
		&\multicolumn{4}{c||}{MSMT17}
		&\multicolumn{4}{c||}{CUHK03}
		&\multicolumn{3}{c}{VIPeR} \\ 
		\cline{2-20}
		&R1 & R5 & R10 & mAP 
		&R1 & R5 & R10 & mAP
		&R1 & R5 & R10 & mAP 
		&R1 & R5 & R10 & mAP
		&R1 & R5 & R10 
		\\
		\hline \hline
		
		Market
		&\cellcolor{lightgray}91.6&\bf\cellcolor{lightgray}97.1&\bf\cellcolor{lightgray}98.0&\bf\cellcolor{lightgray}75.9
		&40.1&54.9&61.2&23.4
		&16.0&24.8&29.4&5.0
		&7.6&15.6&21.1&7.5&29.4&43.4&52.5
		\\
		\hline
		Market+{\bf H}
		&\bf\cellcolor{lightgray}91.8&\cellcolor{lightgray}96.9&\cellcolor{lightgray}97.8&\cellcolor{lightgray}75.6
		&41.3&56.6&62.9&23.9
		&16.3&24.9&29.7&5.1
		&6.6&14.3&20.0&6.8&29.1&43.4&49.4
		\\
		Market+{\bf S}
		
		&\cellcolor{lightgray}91.4&\cellcolor{lightgray}96.8&\bf\cellcolor{lightgray}98.0&\cellcolor{lightgray}75.6
		&40.4&56.1&62.3&23.5
		&16.9&25.9&30.8&5.3
		&7.8&16.8&22.1&7.9&29.4&43.7&53.8\\
		Market+{\bf L}
		&\cellcolor{lightgray}90.9&\cellcolor{lightgray}96.7&\bf\cellcolor{lightgray}98.0&\cellcolor{lightgray}75.1
		&44.8&59.5&65.1&26.5
		&18.9&28.6&33.5&5.9
		&8.4&17.6&22.9&8.4&\bf34.2&47.2&55.4
		\\
		Market+{\bf C}
		
		&\cellcolor{lightgray}91.0&\cellcolor{lightgray}96.7&\cellcolor{lightgray}97.9&\cellcolor{lightgray}74.6
		&43.7&58.3&64.2&25.9
		&21.7&31.8&37.1&7.0
		&9.4&18.9&25.1&9.4&32.3&49.1&56.6\\
		\hline
		Market+{\bf All}
		&\cellcolor{lightgray}90.4&\cellcolor{lightgray}96.5&\cellcolor{lightgray}97.8&\cellcolor{lightgray}73.5
		&\bf46.3&\bf61.2&\bf66.8&\bf26.7
		&\bf24.4&\bf35.7&\bf41.0&\bf7.6
		&\bf10.0&\bf20.1&\bf26.0&\bf9.9&33.2&\bf48.1&\bf58.2\\
		\hline \hline
		Duke
		&57.0&73.9&80.2&30.1
		&\bf83.5\cellcolor{lightgray}&\bf91.8\cellcolor{lightgray}&\bf94.3\cellcolor{lightgray}&\bf70.0\cellcolor{lightgray}
		&24.2&34.9&40.3&8.1
		&8.4&18.2&25.1&8.6&28.2&46.8&57.6
		\\ \hline
		Duke+{\bf H}
		&62.5&78.4&83.6&32.2
		&\cellcolor{lightgray}82.0&\cellcolor{lightgray}91.2&\cellcolor{lightgray}94.0&\cellcolor{lightgray}68.2
		&26.3&37.3&42.7&8.6
		&8.6&17.8&22.9&8.3&30.7&48.7&57.0
		\\
		Duke+{\bf S}
		&60.3&76.8&82.4&31.8
		&\cellcolor{lightgray}82.3&\cellcolor{lightgray}91.2&\cellcolor{lightgray}94.2&\cellcolor{lightgray}68.7
		&26.5&37.6&42.7&8.9
		&8.8&18.8&25.6&9.0&29.4&46.2&55.1
		\\
		Duke+{\bf L}
		&59.6&75.2&80.9&31.1
		&\cellcolor{lightgray}82.4&\cellcolor{lightgray}91.4&\cellcolor{lightgray}94.1&\cellcolor{lightgray}68.5
		&27.1&38.2&43.5&9.2
		&10.6&20.6&26.6&10.4&34.5&53.2&59.5
		\\
		Duke+{\bf C}
		&59.4&74.3&81.1&30.7
		&\cellcolor{lightgray}83.4&\cellcolor{lightgray}91.6&\cellcolor{lightgray}94.1&\cellcolor{lightgray}69.0
		&29.8&41.7&47.3&10.1
		&10.1&20.0&26.0&10.1&33.5&52.2&62.7
		\\
		\hline
		Duke+{\bf All}
		&\bf66.1&\bf81.6&\bf86.3&\bf35.5
		&81.5\cellcolor{lightgray}&90.8\cellcolor{lightgray}&93.1\cellcolor{lightgray}&\cellcolor{lightgray}65.9
		&\bf35.5&\bf48.2&\bf53.7&\bf12.2
		&\bf10.7&\bf21.9&\bf27.5&\bf10.5&\bf35.4&\bf54.1&\bf62.3\\
		\hline 
	\end{tabular}
	
\end{table*}

\vspace{0.2cm}
\noindent{\bf Types of image transformations.}
We examined the contribution of every individual image transformation:
Hue, Saturation, Lightness, and Contrast.
Table \ref{tab:each_transf} shows that
the performance benefits by individual transformations
vary with test domains.
This is reasonable due to the difference 
in the viewing condition characteristics of
distinct domains which typically present
no regularity.
This on the other hand indicates the necessity
of exploiting all the image transformations
for better tackling the domain heterogeneity
during deployment at scale.

\begin{figure} 
	\centering
	\includegraphics[width=1.0\linewidth]{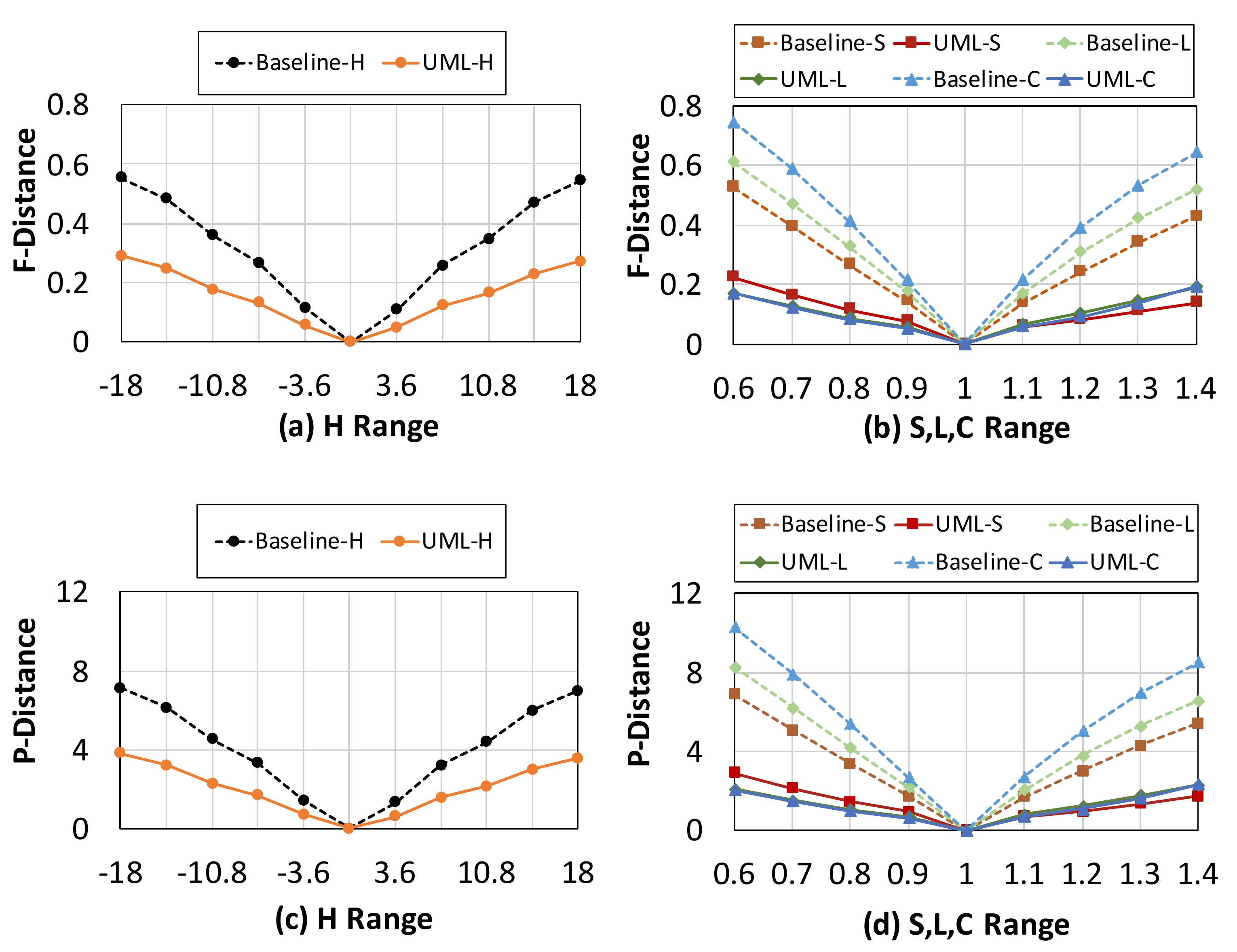} 	
	\
	\caption{Domain universality analysis. 
		Seed domain: DukeMTMC-reID. Test domain: Market-1501.
		F=Feature, P=Prediction,
		H=Hue, S=Saturation, L=Lightness, C=Contrast.}
	\label{fig:domain_univ}
\end{figure}

\vspace{0.2cm}
\noindent{\bf Domain universality.}
We quantified how well the UML model
is generic and universal to various domains 
in the sense of being robust to transformations.
We used the UML model trained with DukeMTMC-reID as the seed domain,
and tested its universality degree on transformed
Market-1501 images.
We selected randomly 1,000 Market-1501 seed (original) images
and applied individual transformations
to each of them. Composited transformations
were not used for simplified and dedicated analysis. 
Such transformations
imitate the cross-domain person appearance variations.
As a comparison, we tested a {\em baseline} model
trained without using our image transformations.

\begin{figure*} 
	\centering
	\includegraphics[width=1.0\linewidth]{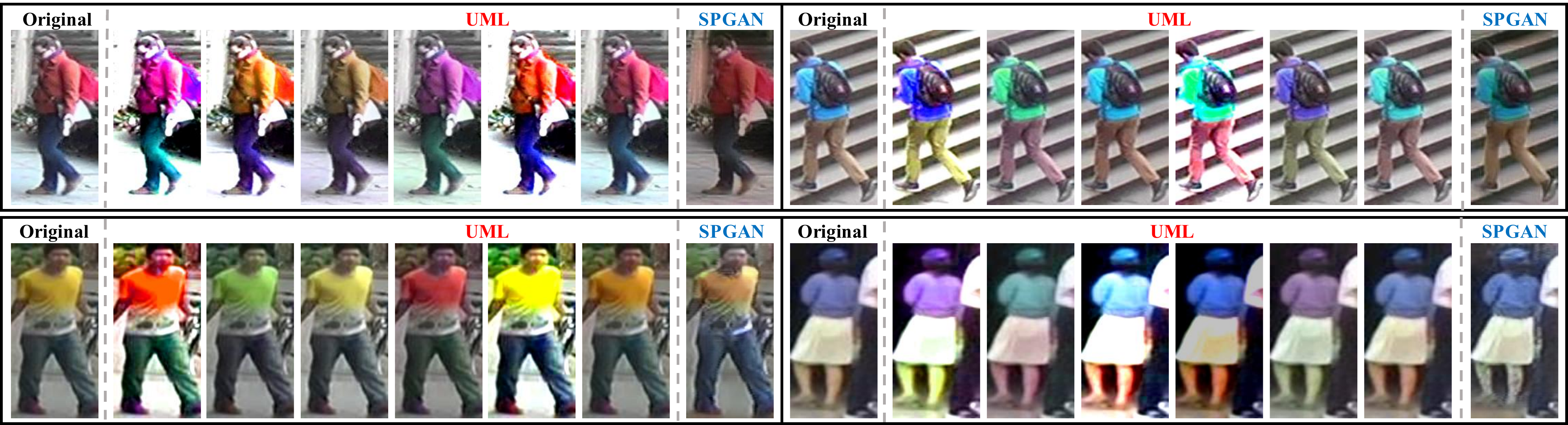} 	
	\caption{
		Visual comparison between UML and SPGAN
		on 
		(top) DukeMTMC-reID and (bottom) Market-1501.
	}
	\label{fig:Visual_GAN_ILR}
\end{figure*}

We considered two measures of domain universality:
(1) {\em Feature level} invariance, and
(2) {\em Prediction level} invariance.
The former is obtained by computing
the Euclidean distance 
of the features of the transformed images
against that of the original images
all extracted by the UML model.
The latter instead is quantified by the Euclidean distance
between their classification prediction vectors.
Figure \ref{fig:domain_univ} shows
that UML enables to learn significantly invariant features
w.r.t. image transformations therefore more robust 
re-id deployment across heterogeneous domains.
This is consistent with the observations
made in Tables \ref{tab:UDA} and \ref{tab:UML}.

\begin{table} 
	\centering
	\setlength{\tabcolsep}{0.22cm}
	\begin{tabular}{l|c|c|c|c|c}
		\hline 
		\multirow{2}{*}{Seed}
		& \multirow{2}{*}{Method}
		& \multicolumn{4}{c}{Duke}\\
		\cline{3-6}
		& & R1 & R5 & R10 & mAP \\
		\hline 
		\multirow{2}{*}{Market} 
		& UML
		& 46.3 & 61.2 & 66.8 & 26.7
		
		\\
		& UML+SPGAN
		&\bf46.9&\bf62.2&\bf67.7&\bf27.0
		\\ 
		\hline \hline
		\multirow{2}{*}{Seed}
		& \multirow{2}{*}{Method}
		& \multicolumn{4}{c}{Market}\\
		\cline{3-6}
		& & R1 & R5 & R10 & mAP \\
		\hline 
		\multirow{2}{*}{Duke} 
		& UML
		& \bf66.1 & \bf81.6 & \bf86.3 & \bf35.5 \\
		& UML+SPGAN& 65.5&81.2&85.8 &35.2
		\\ \hline
	\end{tabular}
	\vspace{0.3cm}
	\caption{
		SPGAN on top of UML.
	}
	\vspace{-0.9cm} 
	\label{tab:spgan_effect}
\end{table}



\begin{figure} [h]
	\centering
	\includegraphics[width=1.0\linewidth]{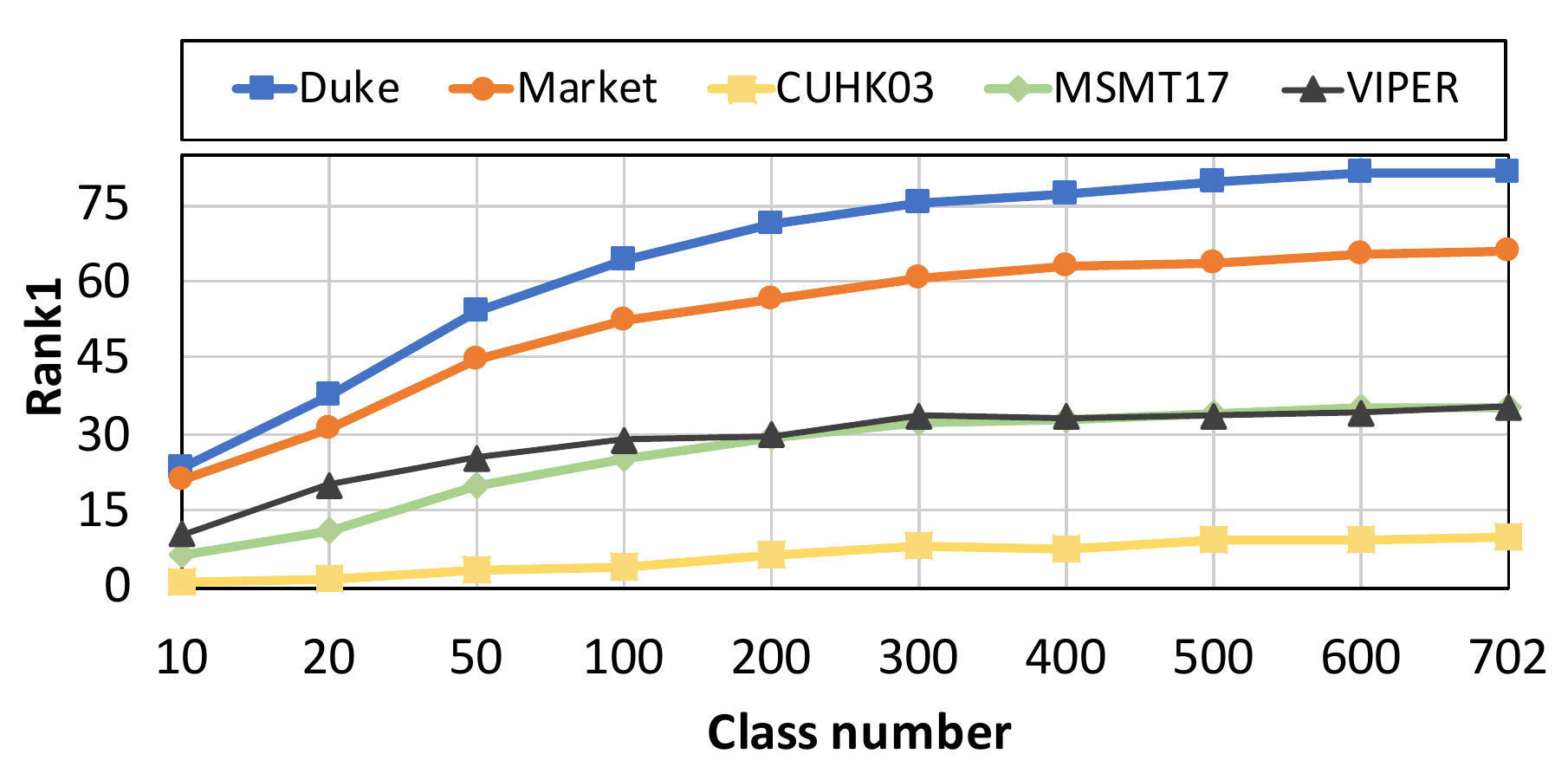} 	
	
	\caption{Effect of seed identity class number. 
		Seed domain: DukeMTMC-reID.}
	\label{fig:ID_num}
	\vspace{-0.2cm}
\end{figure}
\vspace{0.2cm}
\noindent{\bf Comparison to state-of-the-art image synthesis.}
As an image generation method,
we specially compared our UML with the state-of-the-art
image synthesis model SPGAN \cite{deng2018image}.
We did not select HHL \cite{zhong2018generalizing}
since it is a hybrid of image synthesis and cross-domain feature learning.
We have already compared the quantitative results in Table \ref{tab:UDA},
and showed that SPGAN is inferior.
This is intuitively reasonable as observed
from the visual comparison of them in Fig. \ref{fig:Visual_GAN_ILR}.
Specifically, UML generates much more diverse and richer
images than SPGAN in a computationally more efficient
and domain generic manner.
On the contrary, SPGAN requires 
computationally expensive domain-specific model training
along with tedious hyper-parameter tuning.
By only altering the colour and contrast properties,
UML can well preserve the person identity class information
without the need for designing identity preserving loss function.
The colour of clothing and/or associations may 
clearly changes w.r.t. the original seed images,
but all other identity information including person physical and biometric characteristics remain.
This is partly against the conventional understanding
that clothing colour plays the dominant 
role in person re-id therefore their variation 
of the same person identity class may hurt 
the model generalisation \cite{gong2014person}.
Our investigation and finding uniquely
challenge this classical wisdom and 
validate the importance of otherwise appearance information
to person re-id.
This inspires future novel ideas especially for image synthesis modelling. Functionally, SPGAN images can be considered as part of UML images. To demonstrate this, we tested the complementary effect of SPGAN
on top of UML. 
The results in Table \ref{tab:spgan_effect}
show that very limited effect can be resulted 
from adding SPGAN images to the training set.
This also justifies the superior performance of UML
since acquisition of large scale training data is
one of the key elements for ensuring model generalisation
capability in deep learning.

\vspace{0.2cm}
\noindent{\bf Seed identity class number.}
We tested the effect of seed person identity class number
on the model performance.
We used DukeMTMC-reID as the seed training set
and varied the training class number
between 100 and 702.
Figure \ref{fig:ID_num} shows that
more seed classes generally
lead to better performance as expected.
Surprisingly, our method is able to
perform well using as few as 100 person classes
($\frac{1}{7}$ of the standard training size).
This validates the efficacy of our model
in case of limited seed training data.

\section{Conclusion}
\vspace{0.3cm}
In this work we presented a {\em Universal Model Learning} (UML) for domain-generic universal person re-id in a ``train once, run everywhere''
pattern.
This differs from all the existing state-of-the-art
supervised and unsupervised learning (including domain adaptation) methods typically taking 
a ``train once, run once'' pattern,
suffering from per-domain {\em repeated} model training 
as well as the corresponding various costs and limitations.
Our method therefore opens up a direction
taking intelligent learning algorithms closer to industrial-level applications,
although the current performance achieved
is still inferior to
that of supervised learning counterparts.
As a training image generation method, our method
is readily able to integrate any off-the-shelf supervised 
learning algorithms without extra complexity
and obstacle of hyper-parameter tuning
and model optimisation as required 
by image synthesis methods.
Due to such modelling simplicity and flexibility, 
the proposed method serve well as a good baseline upon which further algorithm development and innovation can be established.
We have conducted extensively comparative 
experiments for unsupervised person re-id
in the unlabelled target domain using 
five publicly available benchmarks,
and demonstrated the performance superiority
and modelling advantages of UML over 
the state-of-the-art alternative methods
in both unsupervised model learning
and unsupervised domain adaptation settings.
Detailed component analyses
were also provided for giving insights
and deep understanding about the UML model performance
superiority.

%


\section*{Acknowledgements}
{This work was partly supported by the China Scholarship Council, Vision Semantics Limited, the Royal Society Newton Advanced Fellowship Programme (NA150459), and Innovate UK Industrial Challenge Project on Developing and Commercialising Intelligent Video Analytics Solutions for Public Safety (98111-571149).}

{\small
    \bibliographystyle{aaai}
	\bibliography{rhda}
}

\end{document}